\documentclass[sigconf]{acmart}

\usepackage{amsmath}
\usepackage{amssymb}
\usepackage{amsfonts}
\usepackage{graphicx}
\usepackage{caption}
\usepackage[ruled, linesnumbered,vlined,boxed,commentsnumbered]{algorithm2e}

\usepackage{booktabs} 

\usepackage{bm}
\usepackage{bbm}
\def\BibTeX{{\rm B\kern-.05em{\sc i\kern-.025em b}\kern-.08emT\kern-.1667em\lower.7ex\hbox{E}\kern-.125emX}}

\copyrightyear{2019}
\acmYear{2019} 
\setcopyright{iw3c2w3}
\acmConference[WWW '19 Companion]{Companion Proceedings of the 2019 World Wide Web Conference}{May 13--17, 2019}{San Francisco, CA, USA}
\acmBooktitle{Companion Proceedings of the 2019 World Wide Web Conference (WWW '19 Companion), May 13--17, 2019, San Francisco, CA, USA}
\acmPrice{}
\acmDOI{10.1145/3308560.3316598}
\acmISBN{978-1-4503-6675-5/19/05}

\usepackage{balance}

\begin{document}

%
\title{Reference Product Search}

%

\author{Chu Wang}
\authornote{$\dagger$ Both authors contributed equally.}
\affiliation{Amazon.com}
\email{chuwang@amazon.com}

\author{Lei Tang}
\authornotemark
\affiliation{Amazon.com}
\email{leitang@amazon.com}

\author{Shujun Bian}
\affiliation{Amazon.com}
\email{sjbian@amazon.com}

\author{Da Zhang}
\affiliation{Amazon.com}
\email{dazh@amazon.com}

\author{Zuohua Zhang}
\affiliation{Amazon.com}
\email{zhzhang@amazon.com}

\author{Yongning Wu}
\affiliation{Amazon.com}
\email{yongning@amazon.com}

%
\renewcommand{\shortauthors}{Chu Wang et al.}

%
\begin{abstract}
For a product of interest, we propose a search method to surface a set of reference products. The reference products can be used as candidates to support downstream modeling tasks and business applications. The search method consists of product representation learning and fingerprint-type vector searching. The product catalog information is transformed into a high-quality embedding of low dimensions via a novel attention auto-encoder neural network, and the embedding is further coupled with a binary encoding vector for fast retrieval. We conduct extensive experiments to evaluate the proposed method, and compare it with peer services to demonstrate its advantage in terms of search return rate and precision.
\end{abstract}

%
%



%
\maketitle


\section{Introduction\label{sec:introduction}}


Product modeling and retrieval tasks, including product advertising, substitution, and recommendation, are fundamental for e-commerce business. To help customer discover more products and improve shopping experience, e-commerce platforms like Amazon, eBay, Taobao, and JD all gradually launched new browsing or assistance features related to similar product comparison, alternative product recommendation, and substitute product suggestion. In addition to the traditional query-to-product search, such a product-to-product retrieval mechanism contributes significantly to the e-commerce business. For example, JD launched a ``find similar'' widget available for certain products, and Amazon also provides this widget for products in the browse history, to provide more alternatives closely related to a product of customers' interest. In general, the solutions consist of two stages: a candidate product set is retrieved first, followed by a task-specific ranking model to generate the results. Often, research interests focus on a ranking model built to optimize towards such a business application, but a suitable candidate product set is required to feed the ranking model and it is not well discussed in the literature.


Qualification of the candidate product set differs for different business objectives. In spite of those differences, the candidate product set is generally sourced from catalog information, behavioral data, and human annotations. The traditional inverted-index based retrieval relies on indices which are generated via product attribute tagging, and products sharing common indices like keywords are regarded as the candidates~\cite{gormley2015elasticsearch,smiley2015apache}. If the products are assigned or classified in a taxonomy, the products under the same category can be used as the candidates, though mis-classified query product will lead to irrelevant results, and the size of the retrieved candidate set is uncontrollable and may be highly skewed. Customer behavioral information is another source to generate candidate products. Products that often viewed together or purchased together can serve as candidates for each other~\cite{linden2003amazon}. Though widely adopted in e-commerce business especially for product advertising and recommendation, the mentioned methods need to handle issues like feature sparsity, noisy data sources, low return rate, and inability to adjust the retrieval size. Ideally, the retrieval approach should be able to fetch a enough number of candidate products similar to the query, and the number of candidates should be flexible to adjust in order to accommodate various downstream ranking models for e-commerce applications.


In this paper, we focus on the problem of obtaining a set of reference products for a query product of interest. The reference products can be further fed to a ranking or relevance model to optimize the business objectives like clicks, conversions, or profits. We formulate this product retrieval problem as a search task, where we build product embedding vectors, quantify product similarity by vector distance, and conduct nearest neighbor search (NNS) in the vector space to surface the reference products. In addition to the requirement of flexibly adjusting retrieval size that NNS naturally provides, we focus on two metrics to evaluate the quality of the proposed search method. The precision of the search results measures the quality of the top search results; the return rate estimates the ability to retrieve enough reference products for different queries.


There are various challenges involved to build a desired product-to-product search method. Customer behavior data is limited to only popular products, making it difficult to achieve high return rate. The catalog information is more widely available for product embedding, but feature extraction from low-quality or even missing product information is challenging~\cite{bengio2013representation,xia2014supervised}. In fact, we observe 10\% to 45\% missing rates for several important product fields, and the available catalog information is plagued by poorly written catalog data with irrelevant or duplicated information. Large-scale vector search is also challenging in terms of the computation efficiency. Exact NNS is not realistic for moderately large datasets~\cite{wang2014hashing}, and one need to refer to approximate methods in order to reduce the latency. For datasets with a large volume of items, the balance between the efficiency gain and the quality loss is difficult to handle.
 

In this paper, we propose a novel attention auto-encoder neural network to build high-quality and robust product vectors. In order to achieve better search precision and return rate, the negative impact of missing or low-quality attributes is minimized via attention mechanism~\cite{vaswani2017attention}. The vector search is then conducted by an optimized semantic hashing algorithm~\cite{salakhutdinov2009semantic}, where each product embedding is coupled with a binary encoding so that NNS can be achieved by searching in the vicinity of the query encoding. The encoder is optimized to penalize tiny and huge binary buckets for better search precision and latency. Compared to existing product-to-product retrieval methods, the proposed method is able to obtain better precision while enhancing the return rate drastically. In our experiments for high-traffic products, the proposed method achieves 92.2\% none-zero return rate compared to 86.5\% from the top-performing existing method. This advantage is enlarged to 63.8\% v.s. 3.9\% for general products. As an approximate NNS method, our search method is able to achieve 90.7\% recall rate against the exact 100-NNS compared to 74.2\% or less recall from state-of-the-art packages. As for the computation efficiency, we achieve an average latency of less than 6ms for a product pool of 40 million products with a single machine of 61GB memory and a single Nvidia K80 GPU, and the method can scale up to support large product pools with multiple machines.

\section{Background and Our Contribution\label{sec:previous}}

Product retrieval is fundamental for modeling and business applications in e-commerce. In order for relevance or ranking models to apply, various methods are adopted to retrieve relevant products, including the vastly used inverted-index based methods like Elasticsearch or Solr~\cite{gormley2015elasticsearch,smiley2015apache}. In addition to supporting the downstream business applications, the retrieved products can also be consumed by instance-based product modeling~\cite{aha1991instance,wilson2000reduction}. The retrieval method should be able to fetch enough products for a large proportion of the query products. The retrieval recall rate is also considered a good metric, but it is not commonly adopted because of the lack of ground truth~\cite{schutze2008introduction}. In this paper, the proposed search method is largely related to two areas. Product representation learning and natural language processing help to convert a product to an embedding vector; the approximate nearest neighbor search supports similarity search based on product embeddings at scale. 

Most of the existing product representation techniques for e-commerce depends largely on customer behavior data. Customer co-view and co-purchase information can be directly used to build product embedding such that products with similar browse or purchase history will share similar vectors~\cite{linden2003amazon}. Product catalog data, on the other hand, is a collection of free-form texts like title, description, brand, {\it etc.}. There are supervised and unsupervised ways to transform the catalog data into a vector space. The supervised way originates from ImageNet~\cite{deng2009imagenet}, where a multi-source neural network is trained to predict certain product categorization labels, and the last hidden layer is used as the instance embedding~\cite{simonyan2014very,he2016deep,bengio2013representation}. The categorization labels require extensive annotation while the embedding vector is found of high quality. The unsupervised way, on the other hand, applies to broader cases, especially when labels are unavailable. With the well established methods of word2vec and sentence2vec~\cite{le2014distributed,joulin2016bag,cer2018universal,devlin2018bert}, the problem of product representation learning effectively becomes the task of vector aggregation, namely the method of combining embedding vectors from multiple product fields. Though this problem seems fundamental for natural language processing in e-commerce, to the best of our knowledge, there is no related research dedicated to this problem.

Vector space search has drawn increasing attention because of its ubiquitous applications. In addition to product search, vector search is also frequently used in recommender systems~\cite{koenigstein2012efficient,bachrach2014speeding}, and extreme classifications~\cite{vijayanarasimhan2014deep,weston2011wsabie}. Depending on the definition of the similarity metric, there are nearest neighbor search (NNS) on Euclidean distance, maximum cosine similarity search (MCSS), and maximum inner product search (MIPS)~\cite{shrivastava2014asymmetric,wang2014hashing}. While normalizing the embedding vectors will unify the three, there are independent works for each scenario~\cite{shrivastava2014asymmetric}. The exact search method is not scalable since it requires calculating the pairwise similarity between the query and each of the candidates. Therefore, related works are focusing on approximate methods which can be roughly classified into tree-based approaches, hashing based approaches, and other approaches~\cite{shrivastava2014asymmetric,auvolat2015clustering,hjaltason2003index,seidl1998optimal,jegou2011product}. There are a large amount of works dedicated to high-performance approximate KNN, including FLANN, Iterative Expanding Hashing (IEH), Non-Metric Space Library (NMSLIB), and ANNOY~\cite{gong2013iterative,muja2014scalable,boytsov2013engineering, malkov2018efficient}. Our search method is a fingerprint-type method, belonging to the hashing based approach as IEH, while the other three methods are tree-based approaches.

We discuss technical details of the proposed search method in Section \ref{sec:method}, followed by experiments regarding the return rate, the precision, and the recall according to exact KNN in Section \ref{sec:experiment}. Section \ref{sec:conclusion} discusses possible applications of the proposed method and concludes the paper. Before going into detailed discussions, we would like to highlight our contribution as follows: 
\begin{itemize}
\item
We have developed a product-to-product search method for reference product retrieval. The reference products are obtained via an optimized semantic hashing approach on product embedding generated by a novel attention auto-encoder neural network.
\item
With satisfactory precision, the proposed method is able to achieve considerably higher return rate compared to existing peers. Thus, our reference product search can support general downstream e-commerce applications.
\item
As an approximate KNN method, the proposed search algorithm is able to achieve high recall rate compared to state-of-the-art approximate KNN packages.
\item
It is flexible to adjust the number of returned candidate products based on the requirement of the applications. The latency-precision balance is adjustable in real-time to accommodate different use cases; the product embedding and retrieval encoding are also of a plug-and-play type.
\end{itemize}

\section{Method\label{sec:method}}

In this section, we first describe the offline and online processes of the proposed search method,
followed by detailed discussions on product vectorizer and binary encoder that enable high-quality and fast search for products.

Two transformers are used to convert the text data of product catalog information into vector spaces:
the product vectorizer $g(\cdot)$ and the binary encoder $h(\cdot)$.
For a product $p$, the vectorizer converts the product catalog information into a $d$-dimension embedding (column-)vector 
$\vec{v}=g(p)\in\mathbb{R}^{d}$,
and the binary encoder further transforms the embedding vector into a 
$d'$-dimension binary encoding vector $\vec{b}=h(\vec{v})\in\mathbb{B}^{d'}$.
We will introduce the design and training of the vectorizer $g(\cdot)$ and the encoder $h(\cdot)$ in Section \ref{sec:aae} and \ref{sec:encoder}, respectively.
Let $\mathcal{P}$ denote the product pool where search results are retrieved from, and write $N := |\mathcal{P}|$ the size of $\mathcal{P}$.
Vectorizing and the encoding processes are conducted offline for all the products in $\mathcal{P}$, resulting in an embedding set $V$ and an encoding set $B$. 
For any binary encoding $\vec{b}_i\in B$,  define the corresponding binary encoding bucket $\mathcal{B}_i:=\{\vec{v}\in V \mid h(\vec{v})=\vec{b}_i\}$.
The set of all the buckets $\mathcal{B}^* := \{\mathcal{B}_1, \mathcal{B}_2, \dots\}$ defines a partition over $V$.

%

The online search process follows the semantic hashing mechanism~\cite{salakhutdinov2009semantic}.
We retrieve a small subset consisting of $M$ ($M \ll N$) products from $\mathcal{P}$ via low-cost computation,
followed by applying exact NNS to the candidate set to get the sorted search results.
Such a schema approximates the exact NNS on the larger product pool $\mathcal{P}$ where $M$ is adjusted to balance the quality of approximation and the latency.
For any given query product $p_q$, our reference product search method proceeds as follows:
\begin{enumerate}
\item
Product vectorization.
Calculate the query embedding $\vec{v}_q=g(p_q)$ and the encoding $\vec{b}_q=h(\vec{v}_q)$.
\item
Sort all binary buckets $\mathcal{B}_i$ in $\mathcal{B}^*$ according to the Hamming distance between its encoding $\vec{b}_i$ and the query encoding $\vec{b}_q$ to get a ranked list: $\mathcal{B}_{q_1}, \mathcal{B}_{q_2}, \mathcal{B}_{q_3}, \dots, \mathcal{B}_{q_{|\mathcal{B}^*|}}$.
\item 
Find the minimal cut-off position $c$ such that $|\mathcal{B}_{q_1}| + |\mathcal{B}_{q_2}|+\dots+|\mathcal{B}_{q_c}| \ge M$.
\item
Conduct exact NNS for $\vec{v}_q$ and candidate set $\cup_{i=1}^{c}\mathcal{B}_{q_i}$ and return the top $K$ $(K \le M)$ products.
\end{enumerate}

Note that we conduct exact NNS for the retrieved $M$ products and returns the top-$K$ results as reference products.
In reality, we usually apply a cut-off threshold $\gamma$ to further filter out low quality products to ensure the precision of the search results.
For a specific ranking or relevance model, the search results are further utilized or consumed.
Intuitively, if the retrieved set covers a large proportion of the actual $M$-nearest neighbors of $\vec{v}_q$ in $V$, then the approximation should be good enough for downstream ranking models. We will discuss how to optimize the encoder $h(\cdot)$ for better search quality in Section \ref{sec:encoder}.
Furthermore, we would like to highlight that the product pool can be expanded incrementally without retraining the vectorizer and the encoder. This is advantageous since the time-consuming offline pre-processing is not frequently conducted.

\subsection{Product Representation Learning\label{sec:aae}}
The product vectorizer $g(\cdot)$ converts the product catalog data to a vector
by incorporating signals from multiple attributes.
A single product catalog field is essentially a sentence or a paragraph,
and simple word2vec or sentence2vec model can be applied.
Though there are supervised approaches to obtain product embeddings, 
we choose the unsupervised approach because it can be easily applied to billions of products 
without extensive human annotation efforts.
We train a fastText~\cite{joulin2016bag} model on all the product information (around 400B tokens), 
and use this model as the field vectorizer.
Each product catalog field is then converted into a vector $\vec{u}\in\mathbb{R}^{d}$.

\begin{figure}[htp!]
\centering
\includegraphics[width=0.5\textwidth]{{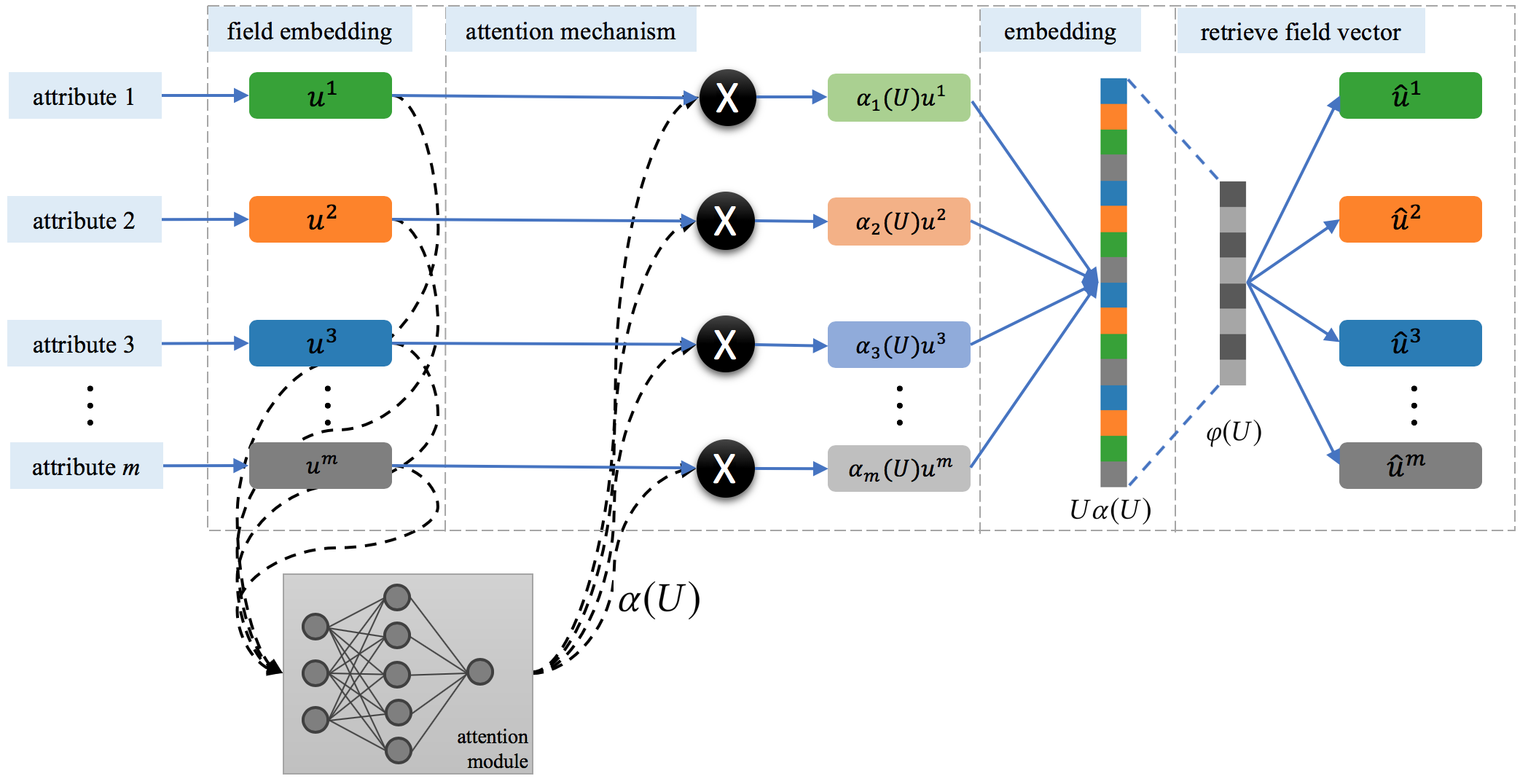}}
\caption{Model structure of the attention auto-encoder.\label{fig:aae}}
\end{figure}

The challenge is how to combine embedding vectors $\vec{u}^1$, $\vec{u}^2$, $\dots$, $\vec{u}^m$ from multiple fields.
Taking more fields into consideration helps utilizing more signals from the product information,
but naively concatenating field vectors suffers from two subsequent problems:
\begin{itemize}
\item
The curse of dimensionality: more fields mean higher dimension of the product embedding,
which increases latency.
\item
Missing fields and low-quality catalog information:
we observe 10\% to 45\% missing rates for several important product fields,
not to mention poorly-written fields with irrelevant or duplicated information.
\end{itemize}
Notice that the data quality issue does not come from data collection and processing. Therefore, it should be alleviated via proper modeling.
To that end, we propose a novel attention auto-encoder network for embedding aggregation.
The goal is that we smartly assign a weight $\alpha_j(U)$ to each field vector $\vec{u}^j\in\mathbb{R}^d$
so that the product embedding is a proper convex combination $\vec{v}_p = g(p) := \sum_{j=1}^m \alpha_j \vec{u}^j$, where the matrix $U := [\vec{u}^1, \vec{u}^2, \dots, \vec{u}^m]$.
The averaging resolves the curse of dimensionality,
and the embedding quality is optimized via minimizing information loss during the averaging.
To be more specific, the weight vector $\vec{\alpha}(U)$ comes from a self-attention module defined as
a softmax distribution:
\begin{equation}\label{eq:softmax}
\vec{\alpha}_j(U) :=\exp(\phi(\vec{u}^j))\Big{/} \sum_{i=1}^m \exp(\phi(\vec{u}^i)),
\end{equation}
where $\phi(\cdot)$ is a simple fully-connected neural network with one hidden layer and a scaler output:
$
\phi(\vec{u}^j) := \vec{\eta}\tanh(W\vec{u}^j+\vec{\xi}).
$
The attention module ``looks at'' each catalog field, 
returns a score estimating the relative amount of information from that field,
and the softmax function further adjusts the weights for balance.
The parameters $W, \vec{\xi}, \vec{\eta}$ are learned via minimizing the information loss from $U$ to $\vec{v}$,
or in other words, to ensure the combined vector $\vec{v}$ can best recover $U$ from another two-layer neural network, defined as
\begin{eqnarray}
\varphi(U) &:=& \sigma\left(W^{(1)}\vec{v}_p+\vec{\xi}^{(1)}\right) = \sigma\left(W^{(1)}U\vec{\alpha}(U)+\vec{\xi}^{(1)}\right),\\
\vec{\hat{u}}^j(U) &:=& W^{(2)}_j\varphi(U) +\vec{\xi}^{(2)},
\end{eqnarray}
where the activation function $\sigma(\cdot)$ is the sigmoid function.
The loss for the neural network is simply the mean squared error:
\begin{equation}
L := \frac{1}{m \times N}\sum_{p\in\mathcal{P}}\sum_{j=1}^{m}\left\|  \vec{u}^j_p - \vec{\hat{u}}_p^j(U) \right\|^2,
\end{equation}
where the loss function $L$ depends on matrices $W, W^{(1)}, W^{(2)}_j (1\le j \le m)$ and vectors $\vec{\xi}, \vec{\xi}^{(1)}, \vec{\xi}^{(2)}, \vec{\eta}$. 
During training, $U$ is generated from the catalog data of 40 million products,
the dimension of the hidden layer is 32 for the attention module, 64 for $\varphi(U)$,
and the model parameters are optimized via Adam Optimizer~\cite{schutze2008introduction}.
Note that for product embedding $g(\cdot)$, we only need the parameters $W, \vec{\xi}, \vec{\eta}$ in the attention module.
The rest of the model parameters are auxiliary to help tune the attention module.
For a given product with field embedding $U$, the attention module gives the optimal weights $\vec{\alpha}(U)$
so that the product embedding $\vec{v}=U\vec{\alpha}(U)$ can best incorporate and recover signals from $U$.
The detailed neural network structure is demonstrated in Figure \ref{fig:aae}.

We conduct extensive experiments to evaluate the quality of embeddings based on the proposed attention auto-encoder (AAE) method.
It is worth noting that the auto-encoder here is irrelevant to the auto-encoder trained for the binary encoding vector $\vec{b}$ in Section \ref{sec:encoder}.
The AAE method outperforms the state-of-the-art embedding methods.
For example, we apply exact K-NNS to retrieve products from an annotated private dataset of 100k products in 1300 categories. 
The performance is measured by precision, which calculates the proportion of retrieved products from the same product category as the query product.
The precision of the exact 3-NNS using the AAE embedding is 0.744,
while the corresponding precisions for vanilla fastText, universal sentence encoder, and Bert
are 0.662, 0.651, and 0.615, respectively~\cite{joulin2016bag,cer2018universal,devlin2018bert}.
We omit more details due to the page limit.
Instead, we choose to conduct experiments directly on the final search method in Section \ref{sec:experiment}.

\subsection{Binary Hashing via Denoising Autoencoder\label{sec:encoder}}
In this subsection, we demonstrate how to obtain the binary encoding vector $\vec{b}=h(\vec{v})$.
The basic idea is to use denoising auto-encoder to compress the embedding $\vec{v}$
into a lower dimension logit vector $\vec{z}$ and find an optimal thresholding vector $\vec{\theta}$ to convert $\vec{z}$ into binaries.
The binary encoding schema originates from Hinton {\it et al.}~\cite{salakhutdinov2009semantic}. 
Our work focuses on optimizing the threshold $\vec{\theta}$ for better retrieval quality and lower latency.
To enhance the search method precision and latency,  the following desired properties of the encoding are proposed:
\begin{enumerate}
\item
The encoding dimension should be low for efficiency. 
\item
Tiny buckets should be avoided, otherwise NNS has to take place on too many buckets and latency issue will come up.
\item
Huge buckets should be avoided for ranking efficiency.
\item
The average bucket size should be controllable for different values of $M$.
\end{enumerate}

Denoising auto-encoders are frequently used for extracting and composing robust features,
where the neural network is fed with manually corrupted data to enhance feature quality and stability~\cite{vincent2008extracting,vincent2010stacked}.
The details of the network structure is well-known to the community, 
and hence we only demonstrate how to optimize the thresholding vector $\vec{\theta}$ in this paper.
Our auto-encoder model consists of a two-layer encoder and a two-layer decoder, 
where the number of the encoded feature layer $d'$ is set to 32 with sigmoid function as the activation.
We use $\vec{z} = [z_1, z_2, \dots, z_{32}]$ to denote the compressed feature from the encoding layer where $z_i\in(0, 1)$ $\forall 1\le i \le 32$, 
and the final binary encoding is obtained via thresholding by $b_i = \mathbbm{1}_{z_i\ge \theta_i}$.

Naively choosing $\theta_i = 0.5$ as in~\cite{salakhutdinov2009semantic} will result in too many buckets with extreme sizes.
Note that now the binary bucket $\mathcal{B}=\mathcal{B}(\vec{\theta})$ depends on the thresholding vector,
and let $n_i(\vec{\theta}) = |\mathcal{B}_i(\vec{\theta})|$ be the bucket size.
We define the following objective function:
\begin{equation}\label{eq:bucket}
L'(\vec{\theta}):=\sum_{i}^{|\mathcal{B}^*|} \left ( \frac{\chi^4}{n_i^2(\vec{\theta})} + n_i^2(\vec{\theta})\right ),
\end{equation}
where $\chi$ is a tunable parameter to balance the average bucket size and the number of buckets.
For a single bucket, the summand in \eqref{eq:bucket} is convex and has a unique minimum at $n_i=\chi$,
and both huge and tiny $n_i$ are naturally penalized.
However, the objective \eqref{eq:bucket} becomes non-convex and even discontinuous with respect to $\vec{\theta}$,
making traditional optimization techniques unsuitable.
To overcome the optimization difficulty, we adopt a continuous genetic algorithm~\cite{anderson2005practical},
with population size 100, mutation rate 0.2, and in each generation 100 pairs are randomly selected for crossover.
We iterate for 200 generations and choose the $\vec{\theta}$ with the smallest objective $L'$ across all the generations.
For our test product pool of 40 million products, the optimized $\vec{\theta}$ for $\chi=100$ is able to increase the average bucket size from 1.78 via the vanilla semantic hashing to 57.32, and the largest bucket size decreases from 60k to 3.5k.
Again, we omit the standalone evaluation of the proposed encoding method, 
and conduct experiments on the overall search method in Section \ref{sec:experiment}.


\section{Experiments\label{sec:experiment}}

%

In this section, we present a series of experiments to evaluate the proposed solution.
We have built a Reference Product Service (RPS) using the AAE product embedding and semantic hashing approximate NNS.
For a given query product, its embedding and binary encoding are computed in real-time, followed by the semantic hashing NNS, which is implemented using a CPU-GPU hybrid model.
We first compare RPS with its peer services and present two quality metrics, 
namely the return rate and the precision at $K$, in Section \ref{sec:coverage} and \ref{sec:precision}, respectively.
Then, as the semantic hashing NNS approximates the Exact NNS, we compare its performance with the Exact NNS in Section \ref{sec:recall}.
Lastly, the complexity and computation efficiency are briefly discussed in Section \ref{sec:latency}.


\subsection{Return Rate Test\label{sec:coverage}}
For a given test pool of queries $\mathcal{Q}_{\mathrm{test}}$, let $\psi(\mathcal{S}, q)$ be the number of search results returned by $\mathcal{S}$ for the query product $q \in \mathcal{Q}_{\mathrm{test}}$. Then the return rate at $K$, denoted by $\mathcal{R}_K(\mathcal{S}, \mathcal{Q}_{\mathrm{test}})$ is defined as
\begin{equation}
\mathcal{R}_K(\mathcal{S}, \mathcal{Q}_{\mathrm{test}}) := \frac{1}{|\mathcal{Q}_{\mathrm{test}}|}\sum_{q \in \mathcal{Q}_{\mathrm{test}}}  \mathbbm{1}_{\psi(\mathcal{S}, q)\ge K},
\end{equation}
where $\mathbbm{1}$ is the indicator function.
The return rate estimates the ability to retrieve enough reference products for different query products.
In reality, this ability is largely limited by product feature availability and the design of the search method.
For example, a search method depending on customer browsing history will not work for a new product without views.
However, recall that one of our goals of the reference products is to make it general so it is able to support different applications.

Two test sets are constructed, namely the purchased set and the general set.
The purchased set consists of $10^6$ randomly sampled products with purchase history.
The general set consists of $10^6$ products randomly sampled from a billion-level product pool.
Products from the purchased set can be considered be higher quality.
We compare the return rate of RPS with four existing product-to-product services.
Two peer services highly depend on behavioral data (denoted by Behavior-1 and Behavior-2),
and the other two depend on general product information (denoted by Method-1 and Method-2).
RPS uses a product pool of $4 \times 10^7$ products with $d = 100$, $d' = 32$ and $M = 4000$.
The related results are listed in Table \ref{table:coverage1} for the purchased query set and Table \ref{table:coverage3} for the general query set. 

\begin {table}
\caption{Return rate at $K$, $\mathcal{R}_K(\mathcal{S}, \mathcal{Q}_{\mathrm{purchased}})$\label{table:coverage1}}
\centering
\begin{tabular}{rrrrrr} 
\toprule
K&	RPS& Method-1&	Method-2&		Behavior-1&	Behavior-2\\
\midrule
1&	0.922&	0.293&	0.865&		0.643&	0.742\\
3&	0.902&	0.293&	0.748&		0.528&	0.688\\
5&	0.893&	0.149&	0.637&		0.475&	0.657\\
10&	0.888&	0.0&	0.408&		0.405&	0.606\\
50&	0.775&	0.0&	0.0&		0.234&	0.415\\
\bottomrule
\end{tabular}
\end{table}
\begin {table}
\caption{Return rate at $K$, $\mathcal{R}_K(\mathcal{S}, \mathcal{Q}_{\mathrm{general}})$\label{table:coverage3}}
\begin{tabular}{rrrrrr} 
\toprule
K&	RPS&	Method-1&	Method-2&	Behavior-1&	Behavior-2\\
\midrule
1&	0.638&	0.008&	0.039&		0.017&	0.030\\
3&	0.599&	0.008&	0.030&		0.013&	0.023\\
5&	0.586&	0.003&	0.024&		0.011&	0.020\\
10&	0.579&	0.00&	0.014&		0.009&	0.017\\
50&	0.456&	0.00&	0.00&		0.004&	0.010\\
\bottomrule
\end{tabular}
\end{table}

We acknowledge that we have no control on the quality or the size of the product pool used by each peer service, nor how the search method is implemented. We simply summarize the observed results as below.
Our service returns enough search results for a larger proportion of query products.
This advantage becomes more obvious for general products where all the peer services fail.
Recall that the return rate estimates the applicability of a solution, better return rates indicate that our solution is able to support broader applications.

\subsection{Precision Test\label{sec:precision}}

The precision at $K$ metric $\mathcal{P}_K$ is defined as the proportion of the top-$K$ results that are indeed similar to the query based on human judgement. 
More specifically, let $\varphi(\mathcal{S}, p, K)$ be the number of positively annotated products from the top-$K$ search results by method $\mathcal{S}$ for a query $q \in Q_\mathrm{test}$, the precision at $K$ metric $\mathcal{P}_K$ is defined as
\begin{equation}
\mathcal{P}_K(\mathcal{S}, \mathcal{Q}_\mathrm{test}) := \frac{1}{|\mathcal{Q}_\mathrm{test}|}\sum_{q \in \mathcal{Q}_\mathrm{test}} \varphi(\mathcal{F}, q, K).
\end{equation}
We observe a sizable proportion of the search results from Behavior-1 and Behavior-2 is dominated by noise since both methods heavily depend on behavioral features,
making them less competitive for the precision test.
Thus, we only annotate the search results from Method-1, Method-2, and our own RPS.
We set $K=5$ and randomly sample 1500 products such that $\psi(\mathcal{S}, q)\ge 5$ for all the three methods.
In order to better demonstrate the difference between the methods,
the annotators are asked to use strict criterion in terms of similarity and to label at least 3 products as negative
out of the 15 products pooled from the top-5 results from each of the three method.
The precision test results are shown in Table \ref{table:quality} 
where RPS outperforms the other two baseline peers.
Note that the results should be viewed for comparison purpose only and the actual precision is higher for all the methods because of the strict criterion.

\begin {table}
\caption{Precision test in each product line\label{table:quality}}
\centering
\begin{tabular}{rrrr} 
\toprule
Product Line	&Method-1	&	Method-2&	RPS\\
\midrule
Softline&	0.582&	0.572&	0.865\\
Hardline&	0.621&	0.656&	0.794\\
Consumable&	0.673&	0.705&	0.827\\
\bottomrule
\end{tabular}
\end{table}

\subsection{Approximate KNN Recall Test\label{sec:recall}}

In addition to the above search quality test, we also conduct the recall test for completeness.
The recall rate measures the proportion of actual positives that are included in the search results,
which is generally not practical since the ground truth is unknown.
Instead, we follow the convention to compute the recall against the exact NNS method and compare
it with four other open-source approximate NNS methods.
Non-Metric Space Library (NMSLIB)~\cite{nmslib}, Approximate Nearest Neighbors Oh Yeah (ANNOY)~\cite{annoyKNN}, and Fast Library for Approximate Nearest Neighbors (FLANN) are tree-based approaches, while Iterative Expanding Hashing is a hashing type approach~\cite{gong2013iterative,muja2014scalable,boytsov2013engineering, malkov2018efficient}.

Since we require the exact KNN results to be the ground truth, 
we use a smaller dataset for the recall test: a set of 2 million products generated randomly.
For each method, we calculate the proportion of the top-$K$ search results that actually hit the top-$K$ products from exact KNN search.
Recall values are presented in Table \ref{table:recall} for $K$ varying from 1 to 100.
Note that the query itself is always excluded from the search results.

\begin {table}
\caption{Recall-at-K test\label{table:recall}}
\centering
\begin{tabular}{rrrrrr} 
\toprule
K	&RPS&	NMSLIB&	FLANN	&	ANNOY	& IEH\\
\midrule
1&	97.24&	90.13&	90.59&	91.35	&	95.14\\
5&	95.82&	93.42&	85.57&	85.94	&	90.88\\
10&	94.85&	92.67&	79.10&	82.62	&	87.65\\
100&	90.72&	61.90&	58.64&	66.90	&	74.21\\
\bottomrule
\end{tabular}
\end{table}

Table \ref{table:recall} shows that our approximate KNN based on binary encoding achieves satisfactory recall rates.
In addition, we would like to highlight its flexibility.
The efficiency-quality parameter $M$ can be adjusted online without preprocessing to balance efficiency and quality,
while most of other approximate KNN algorithms can not.
It also allows easy modification of product pool in realtime. 
For example, adding new embedding and encoding vectors are more easily handled than training a new tree structure.

\subsection{Latency\label{sec:latency}}

In this subsection, we briefly discuss the efficiency and the latency of our search method.
In general, the combined embedding and encoding latency for a single query is under 1ms.
With a single AWS machine of type p2.xlarge (equipped with an NVIDIA K80 GPU), for a product pool of 40 million products and the subset size $M=4000$,
the average search latency is under 6ms, which is in the same scale as the general network latency.
Thus, the proposed method is able to efficiently support general business applications.


\section{Discussion and Conclusion\label{sec:conclusion}}
In this paper, we propose a search method to surface a set of reference products.
The product catalog information is transformed into a high-quality embedding of low dimension via a novel attention auto-encoder
neural network, and the embedding is further coupled with a binary encoding vector for high quality vector search at scale.
We conduct extensive experiments to evaluation the return rate, the precision, and the recall rate of the proposed method,
and compare them with peer methods.
Since our method is able to yield a satisfactory number of high-quality results for most of the query products,
the reference products are readily consumable for various business ranking models to support applications like pricing, substitution, and recommendation.
We believe such an acceleration to support multiple applications will bring fundamental values to the e-commerce business,
and invite more future works on the algorithms and applications of the reference product set.

%
\bibliographystyle{ACM-Reference-Format}
\bibliography{ecnlp}


\begin{thebibliography}{00}


\ifx \showCODEN    \undefined \def \showCODEN     #1{\unskip}     \fi
\ifx \showDOI      \undefined \def \showDOI       #1{#1}\fi
\ifx \showISBNx    \undefined \def \showISBNx     #1{\unskip}     \fi
\ifx \showISBNxiii \undefined \def \showISBNxiii  #1{\unskip}     \fi
\ifx \showISSN     \undefined \def \showISSN      #1{\unskip}     \fi
\ifx \showLCCN     \undefined \def \showLCCN      #1{\unskip}     \fi
\ifx \shownote     \undefined \def \shownote      #1{#1}          \fi
\ifx \showarticletitle \undefined \def \showarticletitle #1{#1}   \fi
\ifx \showURL      \undefined \def \showURL       {\relax}        \fi
\providecommand\bibfield[2]{#2}
\providecommand\bibinfo[2]{#2}
\providecommand\natexlab[1]{#1}
\providecommand\showeprint[2][]{arXiv:#2}

\bibitem[\protect\citeauthoryear{Aha, Kibler, and Albert}{Aha
  et~al\mbox{.}}{1991}]%
        {aha1991instance}
\bibfield{author}{\bibinfo{person}{David~W Aha}, \bibinfo{person}{Dennis
  Kibler}, {and} \bibinfo{person}{Marc~K Albert}.}
  \bibinfo{year}{1991}\natexlab{}.
\newblock \showarticletitle{Instance-based learning algorithms}.
\newblock \bibinfo{journal}{{\em Machine learning\/}} \bibinfo{volume}{6},
  \bibinfo{number}{1} (\bibinfo{year}{1991}), \bibinfo{pages}{37--66}.
\newblock


\bibitem[\protect\citeauthoryear{Anderson-Cook}{Anderson-Cook}{2005}]%
        {anderson2005practical}
\bibfield{author}{\bibinfo{person}{Christine~M Anderson-Cook}.}
  \bibinfo{year}{2005}\natexlab{}.
\newblock \bibinfo{title}{Practical genetic algorithms}.
\newblock   (\bibinfo{year}{2005}).
\newblock


\bibitem[\protect\citeauthoryear{Auvolat, Chandar, Vincent, Larochelle, and
  Bengio}{Auvolat et~al\mbox{.}}{2015}]%
        {auvolat2015clustering}
\bibfield{author}{\bibinfo{person}{Alex Auvolat}, \bibinfo{person}{Sarath
  Chandar}, \bibinfo{person}{Pascal Vincent}, \bibinfo{person}{Hugo
  Larochelle}, {and} \bibinfo{person}{Yoshua Bengio}.}
  \bibinfo{year}{2015}\natexlab{}.
\newblock \showarticletitle{Clustering is efficient for approximate maximum
  inner product search}.
\newblock \bibinfo{journal}{{\em arXiv:1507.05910\/}} (\bibinfo{year}{2015}).
\newblock


\bibitem[\protect\citeauthoryear{Bachrach, Finkelstein, Gilad-Bachrach, Katzir,
  Koenigstein, Nice, and Paquet}{Bachrach et~al\mbox{.}}{2014}]%
        {bachrach2014speeding}
\bibfield{author}{\bibinfo{person}{Yoram Bachrach}, \bibinfo{person}{Yehuda
  Finkelstein}, \bibinfo{person}{Ran Gilad-Bachrach}, \bibinfo{person}{Liran
  Katzir}, \bibinfo{person}{Noam Koenigstein}, \bibinfo{person}{Nir Nice},
  {and} \bibinfo{person}{Ulrich Paquet}.} \bibinfo{year}{2014}\natexlab{}.
\newblock \showarticletitle{Speeding up the xbox recommender system using a
  euclidean transformation for inner-product spaces}. In
  \bibinfo{booktitle}{{\em Proceedings of the 8th ACM Conference on Recommender
  systems}}. ACM, \bibinfo{pages}{257--264}.
\newblock


\bibitem[\protect\citeauthoryear{Bengio, Courville, and Vincent}{Bengio
  et~al\mbox{.}}{2013}]%
        {bengio2013representation}
\bibfield{author}{\bibinfo{person}{Yoshua Bengio}, \bibinfo{person}{Aaron
  Courville}, {and} \bibinfo{person}{Pascal Vincent}.}
  \bibinfo{year}{2013}\natexlab{}.
\newblock \showarticletitle{Representation learning: A review and new
  perspectives}.
\newblock \bibinfo{journal}{{\em IEEE transactions on pattern analysis and
  machine intelligence\/}} \bibinfo{volume}{35}, \bibinfo{number}{8}
  (\bibinfo{year}{2013}), \bibinfo{pages}{1798--1828}.
\newblock


\bibitem[\protect\citeauthoryear{Boytsov and Naidan}{Boytsov and
  Naidan}{2013}]%
        {boytsov2013engineering}
\bibfield{author}{\bibinfo{person}{Leonid Boytsov} {and}
  \bibinfo{person}{Bilegsaikhan Naidan}.} \bibinfo{year}{2013}\natexlab{}.
\newblock \showarticletitle{Engineering efficient and effective non-metric
  space library}. In \bibinfo{booktitle}{{\em International Conference on
  Similarity Search and Applications}}. Springer, \bibinfo{pages}{280--293}.
\newblock


\bibitem[\protect\citeauthoryear{Cer, Yang, Kong, Hua, Limtiaco, John,
  Constant, Guajardo-Cespedes, Yuan, Tar, et~al\mbox{.}}{Cer
  et~al\mbox{.}}{2018}]%
        {cer2018universal}
\bibfield{author}{\bibinfo{person}{Daniel Cer}, \bibinfo{person}{Yinfei Yang},
  \bibinfo{person}{Sheng-yi Kong}, \bibinfo{person}{Nan Hua},
  \bibinfo{person}{Nicole Limtiaco}, \bibinfo{person}{Rhomni~St John},
  \bibinfo{person}{Noah Constant}, \bibinfo{person}{Mario Guajardo-Cespedes},
  \bibinfo{person}{Steve Yuan}, \bibinfo{person}{Chris Tar}, {et~al\mbox{.}}}
  \bibinfo{year}{2018}\natexlab{}.
\newblock \showarticletitle{Universal sentence encoder}.
\newblock \bibinfo{journal}{{\em arXiv:1803.11175\/}} (\bibinfo{year}{2018}).
\newblock


\bibitem[\protect\citeauthoryear{Deng, Dong, Socher, Li, Li, and Fei-Fei}{Deng
  et~al\mbox{.}}{2009}]%
        {deng2009imagenet}
\bibfield{author}{\bibinfo{person}{Jia Deng}, \bibinfo{person}{Wei Dong},
  \bibinfo{person}{Richard Socher}, \bibinfo{person}{Li-Jia Li},
  \bibinfo{person}{Kai Li}, {and} \bibinfo{person}{Li Fei-Fei}.}
  \bibinfo{year}{2009}\natexlab{}.
\newblock \showarticletitle{Imagenet: A large-scale hierarchical image
  database}. In \bibinfo{booktitle}{{\em Computer Vision and Pattern
  Recognition, 2009. CVPR 2009. IEEE Conference on}}. Ieee,
  \bibinfo{pages}{248--255}.
\newblock


\bibitem[\protect\citeauthoryear{Devlin, Chang, Lee, and Toutanova}{Devlin
  et~al\mbox{.}}{2018}]%
        {devlin2018bert}
\bibfield{author}{\bibinfo{person}{Jacob Devlin}, \bibinfo{person}{Ming-Wei
  Chang}, \bibinfo{person}{Kenton Lee}, {and} \bibinfo{person}{Kristina
  Toutanova}.} \bibinfo{year}{2018}\natexlab{}.
\newblock \showarticletitle{Bert: Pre-training of deep bidirectional
  transformers for language understanding}.
\newblock \bibinfo{journal}{{\em arXiv:1810.04805\/}} (\bibinfo{year}{2018}).
\newblock


\bibitem[\protect\citeauthoryear{Erik}{Erik}{2016}]%
        {annoyKNN}
\bibfield{author}{\bibinfo{person}{Bernhardsson Erik}.}
  \bibinfo{year}{2016}\natexlab{}.
\newblock \bibinfo{title}{Approximate Nearest Neigbors OnYeah (Annoy)}.
\newblock \bibinfo{howpublished}{\url{https://github.com/spotify/annoy}}.
  (\bibinfo{year}{2016}).
\newblock


\bibitem[\protect\citeauthoryear{Gong, Lazebnik, Gordo, and Perronnin}{Gong
  et~al\mbox{.}}{2013}]%
        {gong2013iterative}
\bibfield{author}{\bibinfo{person}{Yunchao Gong}, \bibinfo{person}{Svetlana
  Lazebnik}, \bibinfo{person}{Albert Gordo}, {and} \bibinfo{person}{Florent
  Perronnin}.} \bibinfo{year}{2013}\natexlab{}.
\newblock \showarticletitle{Iterative quantization: A procrustean approach to
  learning binary codes for large-scale image retrieval}.
\newblock \bibinfo{journal}{{\em IEEE Transactions on Pattern Analysis and
  Machine Intelligence\/}} \bibinfo{volume}{35}, \bibinfo{number}{12}
  (\bibinfo{year}{2013}), \bibinfo{pages}{2916--2929}.
\newblock


\bibitem[\protect\citeauthoryear{Gormley and Tong}{Gormley and Tong}{2015}]%
        {gormley2015elasticsearch}
\bibfield{author}{\bibinfo{person}{Clinton Gormley} {and}
  \bibinfo{person}{Zachary Tong}.} \bibinfo{year}{2015}\natexlab{}.
\newblock \bibinfo{booktitle}{{\em Elasticsearch: The Definitive Guide: A
  Distributed Real-Time Search and Analytics Engine}}.
\newblock \bibinfo{publisher}{" O'Reilly Media, Inc."}.
\newblock


\bibitem[\protect\citeauthoryear{He, Zhang, Ren, and Sun}{He
  et~al\mbox{.}}{2016}]%
        {he2016deep}
\bibfield{author}{\bibinfo{person}{Kaiming He}, \bibinfo{person}{Xiangyu
  Zhang}, \bibinfo{person}{Shaoqing Ren}, {and} \bibinfo{person}{Jian Sun}.}
  \bibinfo{year}{2016}\natexlab{}.
\newblock \showarticletitle{Deep residual learning for image recognition}. In
  \bibinfo{booktitle}{{\em Proceedings of the IEEE conference on computer
  vision and pattern recognition}}. \bibinfo{pages}{770--778}.
\newblock


\bibitem[\protect\citeauthoryear{Hjaltason and Samet}{Hjaltason and
  Samet}{2003}]%
        {hjaltason2003index}
\bibfield{author}{\bibinfo{person}{Gisli~R Hjaltason} {and}
  \bibinfo{person}{Hanan Samet}.} \bibinfo{year}{2003}\natexlab{}.
\newblock \showarticletitle{Index-driven similarity search in metric spaces
  (survey article)}.
\newblock \bibinfo{journal}{{\em ACM Transactions on Database Systems
  (TODS)\/}} \bibinfo{volume}{28}, \bibinfo{number}{4} (\bibinfo{year}{2003}),
  \bibinfo{pages}{517--580}.
\newblock


\bibitem[\protect\citeauthoryear{Jegou, Douze, and Schmid}{Jegou
  et~al\mbox{.}}{2011}]%
        {jegou2011product}
\bibfield{author}{\bibinfo{person}{Herve Jegou}, \bibinfo{person}{Matthijs
  Douze}, {and} \bibinfo{person}{Cordelia Schmid}.}
  \bibinfo{year}{2011}\natexlab{}.
\newblock \showarticletitle{Product quantization for nearest neighbor search}.
\newblock \bibinfo{journal}{{\em IEEE transactions on pattern analysis and
  machine intelligence\/}} \bibinfo{volume}{33}, \bibinfo{number}{1}
  (\bibinfo{year}{2011}), \bibinfo{pages}{117--128}.
\newblock


\bibitem[\protect\citeauthoryear{Joulin, Grave, Bojanowski, and Mikolov}{Joulin
  et~al\mbox{.}}{2016}]%
        {joulin2016bag}
\bibfield{author}{\bibinfo{person}{Armand Joulin}, \bibinfo{person}{Edouard
  Grave}, \bibinfo{person}{Piotr Bojanowski}, {and} \bibinfo{person}{Tomas
  Mikolov}.} \bibinfo{year}{2016}\natexlab{}.
\newblock \showarticletitle{Bag of tricks for efficient text classification}.
\newblock \bibinfo{journal}{{\em arXiv:1607.01759\/}} (\bibinfo{year}{2016}).
\newblock


\bibitem[\protect\citeauthoryear{Koenigstein, Ram, and Shavitt}{Koenigstein
  et~al\mbox{.}}{2012}]%
        {koenigstein2012efficient}
\bibfield{author}{\bibinfo{person}{Noam Koenigstein},
  \bibinfo{person}{Parikshit Ram}, {and} \bibinfo{person}{Yuval Shavitt}.}
  \bibinfo{year}{2012}\natexlab{}.
\newblock \showarticletitle{Efficient retrieval of recommendations in a matrix
  factorization framework}. In \bibinfo{booktitle}{{\em International
  conference on Information and knowledge management}}. ACM,
  \bibinfo{pages}{535--544}.
\newblock


\bibitem[\protect\citeauthoryear{Le and Mikolov}{Le and Mikolov}{2014}]%
        {le2014distributed}
\bibfield{author}{\bibinfo{person}{Quoc Le} {and} \bibinfo{person}{Tomas
  Mikolov}.} \bibinfo{year}{2014}\natexlab{}.
\newblock \showarticletitle{Distributed representations of sentences and
  documents}. In \bibinfo{booktitle}{{\em International Conference on Machine
  Learning}}. \bibinfo{pages}{1188--1196}.
\newblock


\bibitem[\protect\citeauthoryear{Linden, Smith, and York}{Linden
  et~al\mbox{.}}{2003}]%
        {linden2003amazon}
\bibfield{author}{\bibinfo{person}{Greg Linden}, \bibinfo{person}{Brent Smith},
  {and} \bibinfo{person}{Jeremy York}.} \bibinfo{year}{2003}\natexlab{}.
\newblock \showarticletitle{Amazon. com recommendations: Item-to-item
  collaborative filtering}.
\newblock \bibinfo{journal}{{\em IEEE Internet computing\/}}
  \bibinfo{number}{1} (\bibinfo{year}{2003}), \bibinfo{pages}{76--80}.
\newblock


\bibitem[\protect\citeauthoryear{Malkov and Yashunin}{Malkov and
  Yashunin}{2018}]%
        {malkov2018efficient}
\bibfield{author}{\bibinfo{person}{Yury~A Malkov} {and}
  \bibinfo{person}{Dmitry~A Yashunin}.} \bibinfo{year}{2018}\natexlab{}.
\newblock \showarticletitle{Efficient and robust approximate nearest neighbor
  search using hierarchical navigable small world graphs}.
\newblock \bibinfo{journal}{{\em IEEE transactions on pattern analysis and
  machine intelligence\/}} (\bibinfo{year}{2018}).
\newblock


\bibitem[\protect\citeauthoryear{Muja and Lowe}{Muja and Lowe}{2014}]%
        {muja2014scalable}
\bibfield{author}{\bibinfo{person}{Marius Muja} {and} \bibinfo{person}{David~G
  Lowe}.} \bibinfo{year}{2014}\natexlab{}.
\newblock \showarticletitle{Scalable nearest neighbor algorithms for high
  dimensional data}.
\newblock \bibinfo{journal}{{\em IEEE Transactions on Pattern Analysis \&
  Machine Intelligence\/}} \bibinfo{number}{11} (\bibinfo{year}{2014}),
  \bibinfo{pages}{2227--2240}.
\newblock


\bibitem[\protect\citeauthoryear{Naidan, Boytsov, Yury, David, and Ben}{Naidan
  et~al\mbox{.}}{2016}]%
        {nmslib}
\bibfield{author}{\bibinfo{person}{Bilegsaikhan Naidan},
  \bibinfo{person}{Leonid Boytsov}, \bibinfo{person}{Malkov Yury},
  \bibinfo{person}{Novak David}, {and} \bibinfo{person}{Frederickson Ben}.}
  \bibinfo{year}{2016}\natexlab{}.
\newblock \bibinfo{title}{Non-Metric Space Library (NMSLIB)}.
\newblock \bibinfo{howpublished}{\url{https://github.com/nmslib/nmslib}}.
  (\bibinfo{year}{2016}).
\newblock


\bibitem[\protect\citeauthoryear{Salakhutdinov and Hinton}{Salakhutdinov and
  Hinton}{2009}]%
        {salakhutdinov2009semantic}
\bibfield{author}{\bibinfo{person}{Ruslan Salakhutdinov} {and}
  \bibinfo{person}{Geoffrey Hinton}.} \bibinfo{year}{2009}\natexlab{}.
\newblock \showarticletitle{Semantic hashing}.
\newblock \bibinfo{journal}{{\em International Journal of Approximate
  Reasoning\/}} \bibinfo{volume}{50}, \bibinfo{number}{7}
  (\bibinfo{year}{2009}), \bibinfo{pages}{969--978}.
\newblock


\bibitem[\protect\citeauthoryear{Sch{\"u}tze, Manning, and
  Raghavan}{Sch{\"u}tze et~al\mbox{.}}{2008}]%
        {schutze2008introduction}
\bibfield{author}{\bibinfo{person}{Hinrich Sch{\"u}tze},
  \bibinfo{person}{Christopher~D Manning}, {and} \bibinfo{person}{Prabhakar
  Raghavan}.} \bibinfo{year}{2008}\natexlab{}.
\newblock \bibinfo{booktitle}{{\em Introduction to information retrieval}}.
  Vol.~\bibinfo{volume}{39}.
\newblock \bibinfo{publisher}{Cambridge University Press}.
\newblock


\bibitem[\protect\citeauthoryear{Seidl and Kriegel}{Seidl and Kriegel}{1998}]%
        {seidl1998optimal}
\bibfield{author}{\bibinfo{person}{Thomas Seidl} {and}
  \bibinfo{person}{Hans-Peter Kriegel}.} \bibinfo{year}{1998}\natexlab{}.
\newblock \showarticletitle{Optimal multi-step k-nearest neighbor search}. In
  \bibinfo{booktitle}{{\em ACM Sigmod Record}}, Vol.~\bibinfo{volume}{27}. ACM,
  \bibinfo{pages}{154--165}.
\newblock


\bibitem[\protect\citeauthoryear{Shrivastava and Li}{Shrivastava and
  Li}{2014}]%
        {shrivastava2014asymmetric}
\bibfield{author}{\bibinfo{person}{Anshumali Shrivastava} {and}
  \bibinfo{person}{Ping Li}.} \bibinfo{year}{2014}\natexlab{}.
\newblock \showarticletitle{Asymmetric LSH (ALSH) for sublinear time maximum
  inner product search (MIPS)}. In \bibinfo{booktitle}{{\em Advances in Neural
  Information Processing Systems}}. \bibinfo{pages}{2321--2329}.
\newblock


\bibitem[\protect\citeauthoryear{Simonyan and Zisserman}{Simonyan and
  Zisserman}{2014}]%
        {simonyan2014very}
\bibfield{author}{\bibinfo{person}{Karen Simonyan} {and}
  \bibinfo{person}{Andrew Zisserman}.} \bibinfo{year}{2014}\natexlab{}.
\newblock \showarticletitle{Very deep convolutional networks for large-scale
  image recognition}.
\newblock \bibinfo{journal}{{\em arXiv:1409.1556\/}} (\bibinfo{year}{2014}).
\newblock


\bibitem[\protect\citeauthoryear{Smiley, Pugh, Parisa, and Mitchell}{Smiley
  et~al\mbox{.}}{2015}]%
        {smiley2015apache}
\bibfield{author}{\bibinfo{person}{David Smiley}, \bibinfo{person}{Eric Pugh},
  \bibinfo{person}{Kranti Parisa}, {and} \bibinfo{person}{Matt Mitchell}.}
  \bibinfo{year}{2015}\natexlab{}.
\newblock \bibinfo{booktitle}{{\em Apache Solr enterprise search server}}.
\newblock \bibinfo{publisher}{Packt Publishing Ltd}.
\newblock


\bibitem[\protect\citeauthoryear{Vaswani, Shazeer, Parmar, Uszkoreit, Jones,
  Gomez, Kaiser, and Polosukhin}{Vaswani et~al\mbox{.}}{2017}]%
        {vaswani2017attention}
\bibfield{author}{\bibinfo{person}{Ashish Vaswani}, \bibinfo{person}{Noam
  Shazeer}, \bibinfo{person}{Niki Parmar}, \bibinfo{person}{Jakob Uszkoreit},
  \bibinfo{person}{Llion Jones}, \bibinfo{person}{Aidan~N Gomez},
  \bibinfo{person}{{\L}ukasz Kaiser}, {and} \bibinfo{person}{Illia
  Polosukhin}.} \bibinfo{year}{2017}\natexlab{}.
\newblock \showarticletitle{Attention is all you need}. In
  \bibinfo{booktitle}{{\em Advances in Neural Information Processing Systems}}.
  \bibinfo{pages}{5998--6008}.
\newblock


\bibitem[\protect\citeauthoryear{Vijayanarasimhan, Shlens, Monga, and
  Yagnik}{Vijayanarasimhan et~al\mbox{.}}{2014}]%
        {vijayanarasimhan2014deep}
\bibfield{author}{\bibinfo{person}{Sudheendra Vijayanarasimhan},
  \bibinfo{person}{Jonathon Shlens}, \bibinfo{person}{Rajat Monga}, {and}
  \bibinfo{person}{Jay Yagnik}.} \bibinfo{year}{2014}\natexlab{}.
\newblock \showarticletitle{Deep networks with large output spaces}.
\newblock \bibinfo{journal}{{\em arXiv:1412.7479\/}} (\bibinfo{year}{2014}).
\newblock


\bibitem[\protect\citeauthoryear{Vincent, Larochelle, Bengio, and
  Manzagol}{Vincent et~al\mbox{.}}{2008}]%
        {vincent2008extracting}
\bibfield{author}{\bibinfo{person}{Pascal Vincent}, \bibinfo{person}{Hugo
  Larochelle}, \bibinfo{person}{Yoshua Bengio}, {and}
  \bibinfo{person}{Pierre-Antoine Manzagol}.} \bibinfo{year}{2008}\natexlab{}.
\newblock \showarticletitle{Extracting and composing robust features with
  denoising autoencoders}. In \bibinfo{booktitle}{{\em Proceedings of the 25th
  international conference on Machine learning}}. ACM,
  \bibinfo{pages}{1096--1103}.
\newblock


\bibitem[\protect\citeauthoryear{Vincent, Larochelle, Lajoie, Bengio, and
  Manzagol}{Vincent et~al\mbox{.}}{2010}]%
        {vincent2010stacked}
\bibfield{author}{\bibinfo{person}{Pascal Vincent}, \bibinfo{person}{Hugo
  Larochelle}, \bibinfo{person}{Isabelle Lajoie}, \bibinfo{person}{Yoshua
  Bengio}, {and} \bibinfo{person}{Pierre-Antoine Manzagol}.}
  \bibinfo{year}{2010}\natexlab{}.
\newblock \showarticletitle{Stacked denoising autoencoders: Learning useful
  representations in a deep network with a local denoising criterion}.
\newblock \bibinfo{journal}{{\em Journal of machine learning research\/}}
  \bibinfo{volume}{11}, \bibinfo{number}{Dec} (\bibinfo{year}{2010}),
  \bibinfo{pages}{3371--3408}.
\newblock


\bibitem[\protect\citeauthoryear{Wang, Shen, Song, and Ji}{Wang
  et~al\mbox{.}}{2014}]%
        {wang2014hashing}
\bibfield{author}{\bibinfo{person}{Jingdong Wang}, \bibinfo{person}{Heng~Tao
  Shen}, \bibinfo{person}{Jingkuan Song}, {and} \bibinfo{person}{Jianqiu Ji}.}
  \bibinfo{year}{2014}\natexlab{}.
\newblock \showarticletitle{Hashing for similarity search: A survey}.
\newblock \bibinfo{journal}{{\em arXiv:1408.2927\/}} (\bibinfo{year}{2014}).
\newblock


\bibitem[\protect\citeauthoryear{Weston, Bengio, and Usunier}{Weston
  et~al\mbox{.}}{}]%
        {weston2011wsabie}
\bibfield{author}{\bibinfo{person}{Jason Weston}, \bibinfo{person}{Samy
  Bengio}, {and} \bibinfo{person}{Nicolas Usunier}.}
\newblock \showarticletitle{Wsabie: Scaling up to large vocabulary image
  annotation}.
\newblock


\bibitem[\protect\citeauthoryear{Wilson and Martinez}{Wilson and
  Martinez}{2000}]%
        {wilson2000reduction}
\bibfield{author}{\bibinfo{person}{D~Randall Wilson} {and}
  \bibinfo{person}{Tony~R Martinez}.} \bibinfo{year}{2000}\natexlab{}.
\newblock \showarticletitle{Reduction techniques for instance-based learning
  algorithms}.
\newblock \bibinfo{journal}{{\em Machine learning\/}} \bibinfo{volume}{38},
  \bibinfo{number}{3} (\bibinfo{year}{2000}), \bibinfo{pages}{257--286}.
\newblock


\bibitem[\protect\citeauthoryear{Xia, Pan, Lai, Liu, and Yan}{Xia
  et~al\mbox{.}}{2014}]%
        {xia2014supervised}
\bibfield{author}{\bibinfo{person}{Rongkai Xia}, \bibinfo{person}{Yan Pan},
  \bibinfo{person}{Hanjiang Lai}, \bibinfo{person}{Cong Liu}, {and}
  \bibinfo{person}{Shuicheng Yan}.} \bibinfo{year}{2014}\natexlab{}.
\newblock \showarticletitle{Supervised hashing for image retrieval via image
  representation learning.}
\newblock


\end{thebibliography}

%

\end{document}